\documentclass[12pt]{article}

\usepackage{amssymb}

\usepackage{graphicx}

\date{}

\begin{document}

\title{Canonical Pose Reconstruction from Single Depth Image for 3D Non-rigid Pose Recovery on Limited Datasets}

\author{Fahd Alhamazani$^1$, Yu-Kun Lai$^2$, Paul L. Rosin$^2$\\
\\
$^1$Northern Border University, Rafha, KSA.\\
\tt{email: fahd.alhamzani@nbu.edu.sa}\\
\\
$^2$Cardiff University, Cardiff, UK. \\
\tt{email: \{laiy4, rosinpl\}@cardiff.ac.uk}
}

\maketitle

\begin{abstract}
3D reconstruction from 2D inputs, especially for non-rigid objects like humans, presents unique challenges due to the significant range of possible deformations. Traditional methods often struggle with non-rigid shapes, which require extensive training data to cover the entire deformation space. This study addresses these limitations by proposing a canonical pose reconstruction model that transforms single-view depth images of deformable shapes into a canonical form. This alignment facilitates shape reconstruction by enabling the application of rigid object reconstruction techniques, and supports recovering the input pose in voxel representation as part of the reconstruction task, utilizing both the original and deformed depth images. Notably, our model achieves effective results with only a small dataset of approximately 300 samples. Experimental results on animal and human datasets demonstrate that our model outperforms other state-of-the-art methods.
\end{abstract}

\section{Introduction}
3D reconstruction aims to turn 2D input such as images into 3D shapes. Most 3D reconstruction methods are designed for rigid shapes. But for non-rigid objects that can bend or twist, such as humans or animals, it gets tricky. These objects can have a large range of deformation, making them hard to handle especially for reconstruction tasks, as significant training examples are required to cover the deformation space. To make the problem more manageable, an effective approach is to bring non-rigid shapes back to a default or standardised pose. This default pose is called the canonical form. Using this form can help simplify and improve various geometric processing tasks, from shape retrieval to shape reconstruction.

Canonical form refers to a normalised representation of a deformable shape such that various instances of similar objects are represented in a unified pose which removes the non-rigid deformation. This uniform representation aids in reducing variability \cite{bustos2014coulomb}, ensuring consistency, and simplifying subsequent computational processes \cite{haj2018local}. The canonical form is commonly used in retrieval tasks, enabling us to search for and identify similar 3D models regardless of their deformations. 
However, current canonical form methods often prioritise discriminating between shapes but  fail to retain good quality of shape appearance. These approaches typically rely on either Euclidean distance \cite{pickup2015euclidean,lian2013feature}  or geodesic distance \cite{elad2003bending} which can distort the deformed shapes. Alternatively, some works suggest other approaches like mapping the deformed shape to a template to preserve shape appearance \cite{lian2013feature}. However, these methods all assume that the input is a complete deformed shape, so cannot be applied to cases with depth image input. 

In terms of non-rigid shape completion, unlike existing methods \cite{yan2022shapeformer,zhou2023human,chibane2020implicit} that rely heavily on large-scale datasets to achieve accurate 3D reconstruction, our model demonstrates effective non-rigid 
shape reconstruction from a single depth image, utilising a considerably smaller dataset. This approach addresses the challenge of data efficiency in 3D reconstruction and highlights the model’s capability to generalise accurately even with limited training samples.

In this study, we address the problem of transforming a deformable shape, represented as a single-view depth image, into its canonical form as the \textbf{first stage} of our approach. This initial step is particularly challenging, as the input does not represent a complete shape. In the \textbf{second stage}, we utilise the reconstructed canonical pose to estimate the full 3D shape of the non-rigid object.

\begin{figure*}[t]
    \centering
    \includegraphics[width=\textwidth]{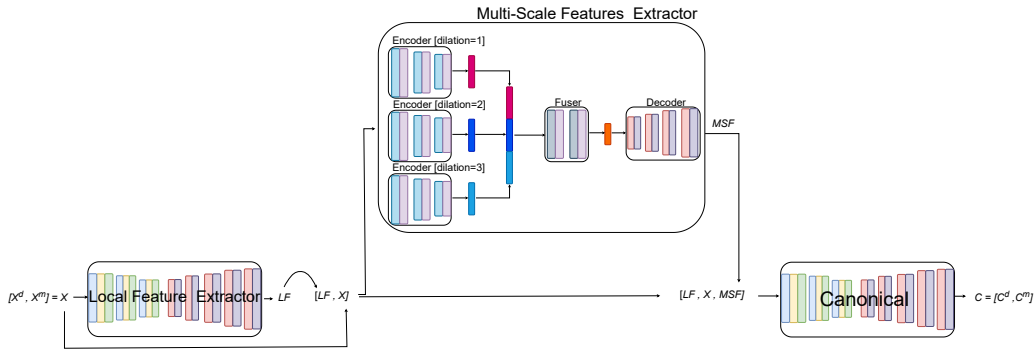} 
    \caption{Stage one, overview of model. where the input depth image in any pose and output canonicalised depth image}
    \label{fig:staage_one_overview}
\end{figure*}

To address this challenging task, we
introduce a learning-based model that turns a single depth image to a default pose. Given a 2D depth image and its corresponding mask, our model aims to produce a depth image that corresponds to the input shape in a canonical pose. Figure \ref{fig:staage_one_overview} displays an overview of the model for the first stage (canonical pose depth estimation), which begins with an encoder-decoder that produces high-dimensional local features. Additionally, we introduce parallel encoders utilising sparse convolution to detect  neighbours sizes, thereby fusing multi-scale features that contribute to preserving shape appearance. These fused features serve as a basis to generate high-dimensional attributes. Ultimately, we use an encoder-decoder model to reconstruct the canonical pose depth image.

In Stage Two, our model incorporates both a pose encoder and a shape encoder to estimate the 3D shape from the canonical pose depth image and the original input depth image. The pose encoder processes the original input depth image, while the shape encoder processes the canonical depth image to capture structural details. Finally, we employ a discriminator, where GAN divergence is used to smooth the volume surface and refine 3D reconstruction quality. This setting allows the model to recover the complete 3D shape with high fidelity.

Our contributions are:
\begin{itemize}

 \item 
  We propose a canonical pose reconstruction model, an end-to-end 2D network designed for the canonical pose reconstruction of single-view depth images. It comprises three components, Local Features Extractor (LF), Multi-Scale Features Extractor (MSF) and canonical pose reconstruction component.
  
  \item 
 We propose parallel encoders and a single decoder block that extract features at different scales and use a fusing decoder to decode multi-scale, high-dimensional features.
  \item 
  We propose a model that estimates the full 3D shape with recovered pose.
  \item 
 The extensive experimental results on TOSCA~\cite{bronstein2008numerical} and human~\cite{pickup2016shape} datasets demonstrate that our model outperforms the existing state-of-the-art methods. Moreover, our model is also capable of preserving high quality shape details while deforming shapes across different types of forms, such as humans and animals with limited datasets.

\end{itemize}
\section{Related Work}
Many tasks including 3D reconstruction and shape retrieval benefit from putting deformable shapes into some standardised poses (such as T-pose for human bodies), which are referred to as canonical forms.
For example, 
shape retrieval is an important task that aims to find similar shapes to the query. 
Many methods work well on rigid bodies where all shapes have fixed pose. However, these methods may work poorly on non-rigid shapes, where the same shape can have different poses. Without a standardised pose (canonical form), determining correspondence between points on two non-rigid shapes can be ambiguous, as the geometric distances caused by pose difference are often much larger than those of different instances. Also, machine learning algorithms, especially those based on deep learning, require consistent data representation for effective training, such as learning-based 3D reconstruction. Different poses can be seen as ``noise'' or ``variations'' that can affect the learning process if not standardised through canonical forms. To solve that, a canonical forms standardises the shapes to a fixed pose. In this section we will review canonical form for non-rigid shapes techniques.

Canonical form reconstruction for non-rigid shapes has been a hot area of research due to its utility in tasks like shape matching and retrieval. Early approaches often relied on geometrically motivated transformations. \cite{lian2013feature} introduced a feature-preserving canonical form using Multidimensional Scaling (MDS) to transform non-rigid 3D watertight meshes. Their method segments objects into near-rigid parts and optimises alignment, preserving key features by minimising non-linear deformations. Although effective, this method can be computationally intensive, limiting its scalability to larger datasets. In contrast, \cite{pickup2015euclidean} achieved computational efficiency by using Euclidean distances between selected vertices to approximate global geodesic distances, offering a faster alternative suitable for high-resolution meshes. However, this approach is limited in adaptability, as it depends on vertex conformal factors to identify structural features, which can be insufficient for highly deformable shapes. Other efforts, like the work of \cite{lian2010non}, approached shape matching with image-based methods, using MDS and PCA to capture the canonical pose of objects in multi-view depth images. Though computationally lighter, these multi-view representations require extensive feature extraction, adding complexity to real-time applications.

Another major stream of works have focused on embedding techniques and multi-feature fusion to handle more complex, non-rigid deformations. \cite{wang2012contour} introduced the contour canonical form, leveraging geodesic constraints to ensure isometry invariance but at a cost of increased geodesic calculations. Building on the need for more flexible approaches, \cite{zeng2019multi} combined canonical forms with multi-view convolutional neural networks, enhancing feature extraction through multi-feature fusion methods, though heavily reliant on extensive data. Meanwhile, \cite{jribi2015novel} proposed a method using geodesic distances to reference points, addressing the challenge of inelastic deformations. More recent works by \cite{haj2018local, haj2017local} used random walks and local commute time distance, allowing shape retrieval by segmenting objects into localised regions, which are then merged for pose-invariant canonical forms. Although these local methods improve retrieval accuracy by preserving salient features, they can struggle to maintain global shape coherence, underscoring the need for approaches that balance global structure with local detail retention.
\section{Methodology}

The canonical form involves addressing deformation by eliminating it, aiming to transform the input depth image (with values ranging from $0$ to $1$) to align with a standardized canonical form. This model comprises four main components. First, the initial component interprets the depth image to extract high-dimensional local features, which are then integrated with the original depth information (Section~\ref{section_lfe}). Next, the model employs parallel encoders alongside a fusing decoder to generate multi-scale features (Section~\ref{section_msfe}). These multi-scale features are subsequently concatenated with the local features, creating skip connections for the canonical pose reconstruction component and grouping local features with multi-scale features to assist in reconstructing the depth image into its canonical form (Section~\ref{sec:deformation}). Finally, we combine the original depth image with the estimated canonical depth image to reconstruct the 3D shape with pose recovery (Section~\ref{poser_recovery}).
\begin{figure}[h]
\centering
\includegraphics[width=\textwidth]{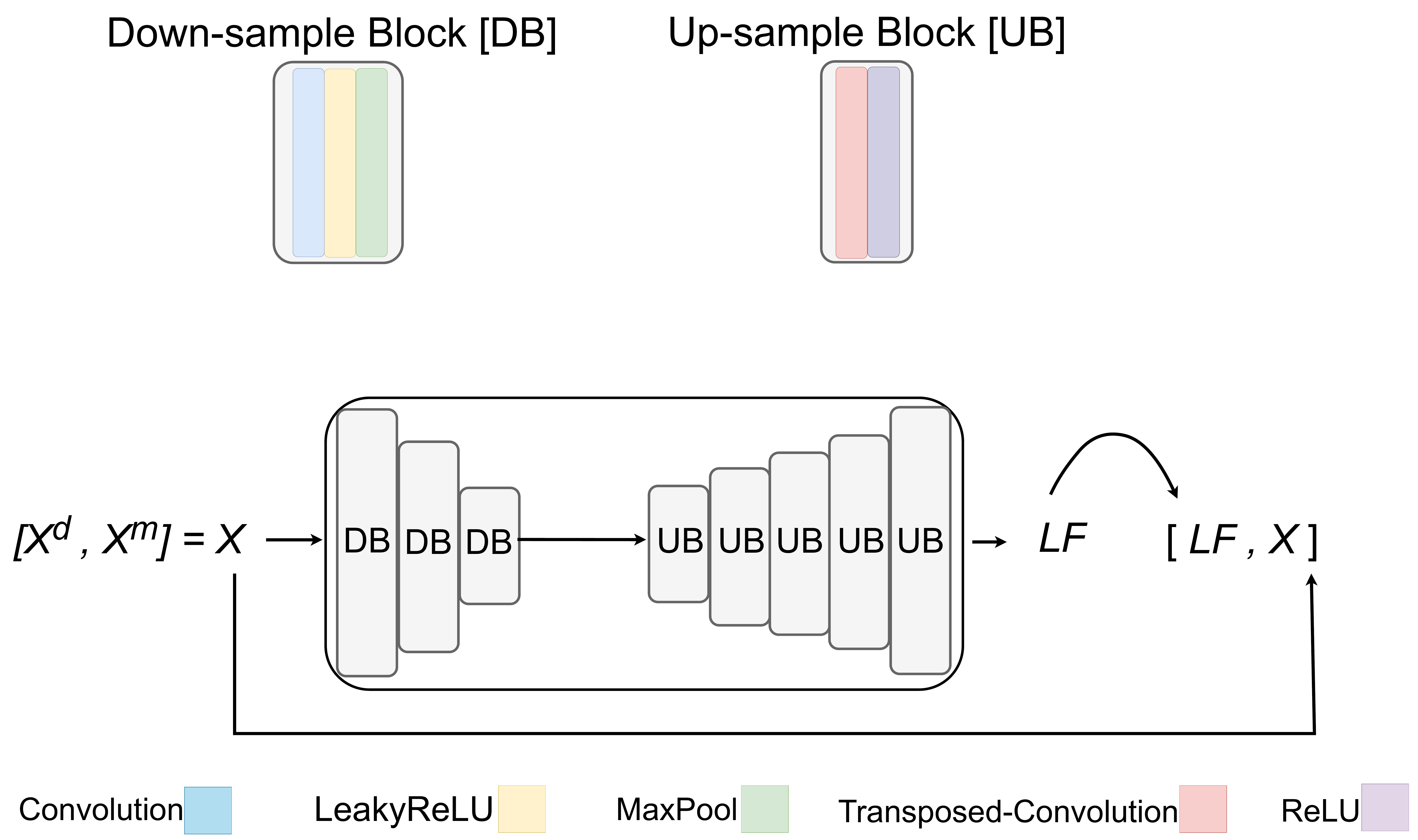} 
\caption{The Local Feature Extractor (LFE) takes the single-view depth image $X^d$ and the corresponding mask $X^m$ as input 
and produces a local feature output, of the same input size, denoted as $Y_{LFE}$}
\label{fig:lfe}

\end{figure}
\subsection{Local Feature Extractor}
\label{section_lfe}
Given an input depth image $X^d$ and mask $X^m$, where $X^{d,m}=\{x^{d,m}_i \in \mathbb{R}^{500 \times 500}\}$, the LFE component processes both to generate local features. The component consists of \(N\) down-sample blocks and \(K\) up-sample blocks, where \(N=3\) and \(K=5\). For the down-sample blocks, each block consists of a convolution with a kernel size of \(5 \times 5\) and strides of \(1 \times 1\). We use \texttt{LeakyReLU} as the activation function, and a \texttt{Maxpool} layer is employed for spatial reduction. For the up-sample blocks, the transpose-convolutions use three different kernel sizes: \([5,3,2]\). The output features, denoted as $Y_{LF}$ in Eq. \ref{eq_lfe}, are concatenated with the original input \(X^{d,m}\) as an extra channel. The network is shown in Figure \ref{fig:lfe}. 
\begin{equation}
 LF = LFE(X^d,X^m)
\label{eq_lfe}
\end{equation}
LFE attaches local features to the original depth image and its mask, so each pixel is associated with both a local feature and a mask value. Consequently, in Section~\ref{sec:deformation}, the reconstruction component has access to both the local features and the original input depth image.

\subsection{Multi-Scale  Feature Extractor}
\label{section_msfe}
The Multi-scale Feature Extractor (MSF) described in Eq. \ref{eq_msfe} comprises three parallel encoders $E_{dilation1}$, $E_{dilation2}$ and $E_{dilation3}$. 
\begin{equation}
 MSF = MSFE(X^d, X^m, LF)
\label{eq_msfe}
\end{equation}
\begin{equation}
 z_{1} = E_{dilation1}(X^d,X^m , LF)
\label{eq_msfe_e1}
\end{equation}
\begin{equation}
 z_{2} = E_{dilation2}(X^d,X^m , LF)
\label{eq_msfe_e2}
\end{equation}
\begin{equation}
 z_{3} = E_{dilation3}(X^d,X^m , LF)
\label{eq_msfe_e3}
\end{equation}
Each encoder captures a different spatial neighbourhood size owing to the inherent nature of convolutions with distinct dilation values. Specifically, the three encoders possess dilation values of 1, 2, and 3 (Eqs. \ref{eq_msfe_e1}, \ref{eq_msfe_e2} and \ref{eq_msfe_e3}) in their convolution layers. {We could not add more than 3 encoders as computation consumption exceeds GPU limits, also 1,2,3 variation is a natural way to expand.}
Every encoder outputs a latent code of size 1600 ($dim(z_{1})=1600$, $dim(z_{2})=1600$ and $dim(z_{3})=1600$, where $dim(\cdot)$ is the dimension of the latent code). {We found that using less than 1600 for the latent code degrades the reconstruction results.} When concatenated, this results in a latent code with a length of 4800.

\begin{figure}[ht]
\centering
\includegraphics[width=\textwidth]{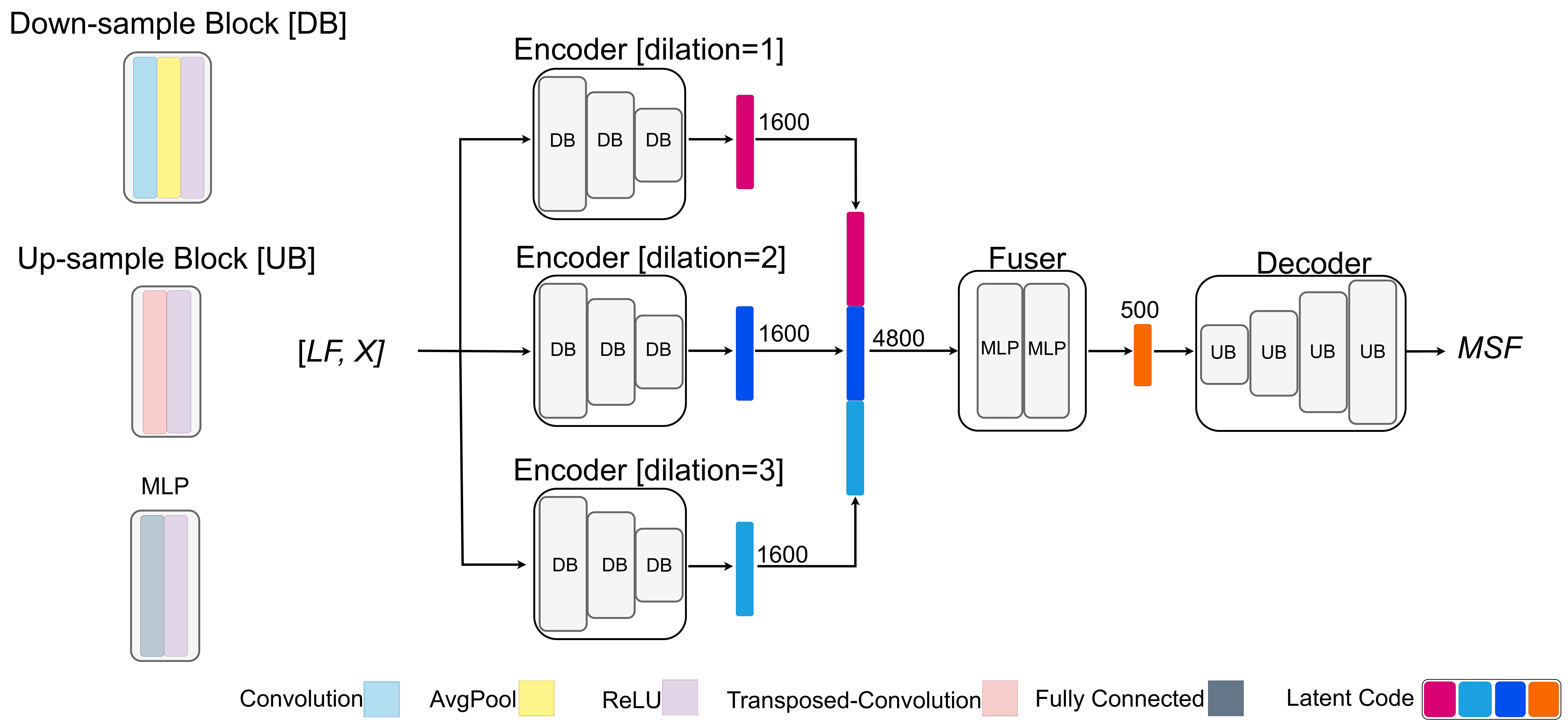} 
\caption{The model takes as input the original depth \(X^d\), its mask \(X^m\) where \([X^d, X^m]=X\), and the local feature output \(Y_{LFE}\). It features three encoders, each having a distinct dilation rate, with each encoder made up of down-sample blocks. Following the encoders, the latent codes are concatenated and passed through a fuser for inter-mapping. The subsequent decoder consists of up-sample blocks, culminating in the reconstructed multi-scale features, denoted as \(MSF\)}
\label{fig:msfe}

\end{figure}
These parallel encoders handle pixels from different scales, thereby yielding multi-scale features. To fuse these multi-scale features, we use a single decoder, as described in Eq. \ref{eq_msfe_d}. 
\begin{equation}
 MSF = D_{fuser}(z_{1},z_{2},z_{3})
\label{eq_msfe_d}
\end{equation}
As an initial step, the 4800D latent codes are processed through two MLP layers to identify inter-code relationships, ultimately generating 500D latent codes. Subsequently, six transpose-convolutions are applied. Following each convolution, a \texttt{ReLU} activation function is employed. The output $MSF$ has the spatial resolution aligned with the original input size. This design enables the association of multi-scale features with each input pixel. The overview of \textit{MSF} is shown in figure \ref{fig:msfe}.

\subsection{Canonical Pose Depth Reconstruction}\label{sec:deformation}
Deformation involves transforming a shape from any pose to a default pose. In terms of an image, this means shifting the pixels to recreate a canonical pose. However, conventional convolution cannot adequately attend to long dependencies. As a solution, we generate both local and multi-scale features of the same size as the input image, allowing the reconstruction component in Eq. \ref{eq_reconstruction} to access both feature types for each pixel.

\begin{figure}[h]
\centering
\includegraphics[width=\textwidth]{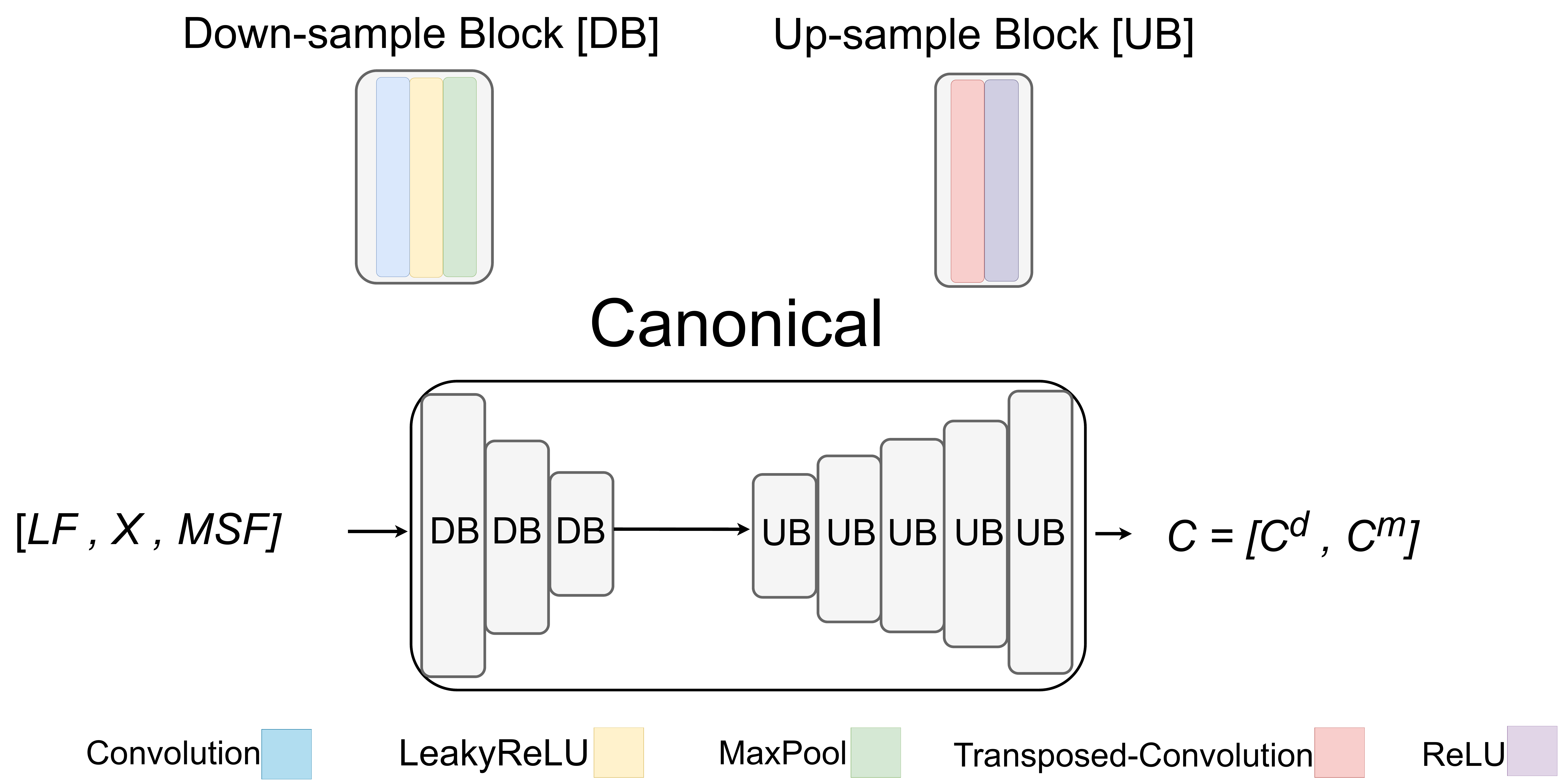} 
\caption{The canonical reconstruction component leverages the original input \(X\), the LFE output \(LF\), and the MSFE output \(MSF\). The model uses these inputs to determine the canonical form $C$ which consists of canonical form depth image $C^d$ and its mask $C^m$.}
\label{fig:deformation}
\end{figure}

Similar to the LF component, the reconstruction component incorporates four channels: the original input and its mask, local feature data generated by the LF component, and multi-scale features produced by MSF. {Note that combining features from different stages of the model helps reduce vanishing gradients}. The reconstruction component comprises \(N\) down-sample blocks and \(K\) up-sample blocks, where \(N=3\) and \(K=5\). Each down-sample block consists of a convolution layer, followed by a \texttt{LeakyReLU} and a \texttt{Maxpool} layer, with kernels of size 5 and stride 1. On the other hand, each up-sample block features a transpose-convolution and a \texttt{ReLU} layer. The final output from the canonical pose component $C^{d,m}$ is a reconstructed depth image alongside a reconstructed mask, where $C^{d,m} \in \mathbb{R}^{500 \times 500}$. The overview of the reconstruction component is shown in figure~\ref{fig:deformation}.

\begin{equation}
 C^{d,m} = canonical(X^d,X^m , LF,MSF)
\label{eq_reconstruction}
\end{equation}

\subsection{Pose Recovery}  
\label{poser_recovery}
\begin{figure}[h]
\centering
\includegraphics[width=\textwidth]{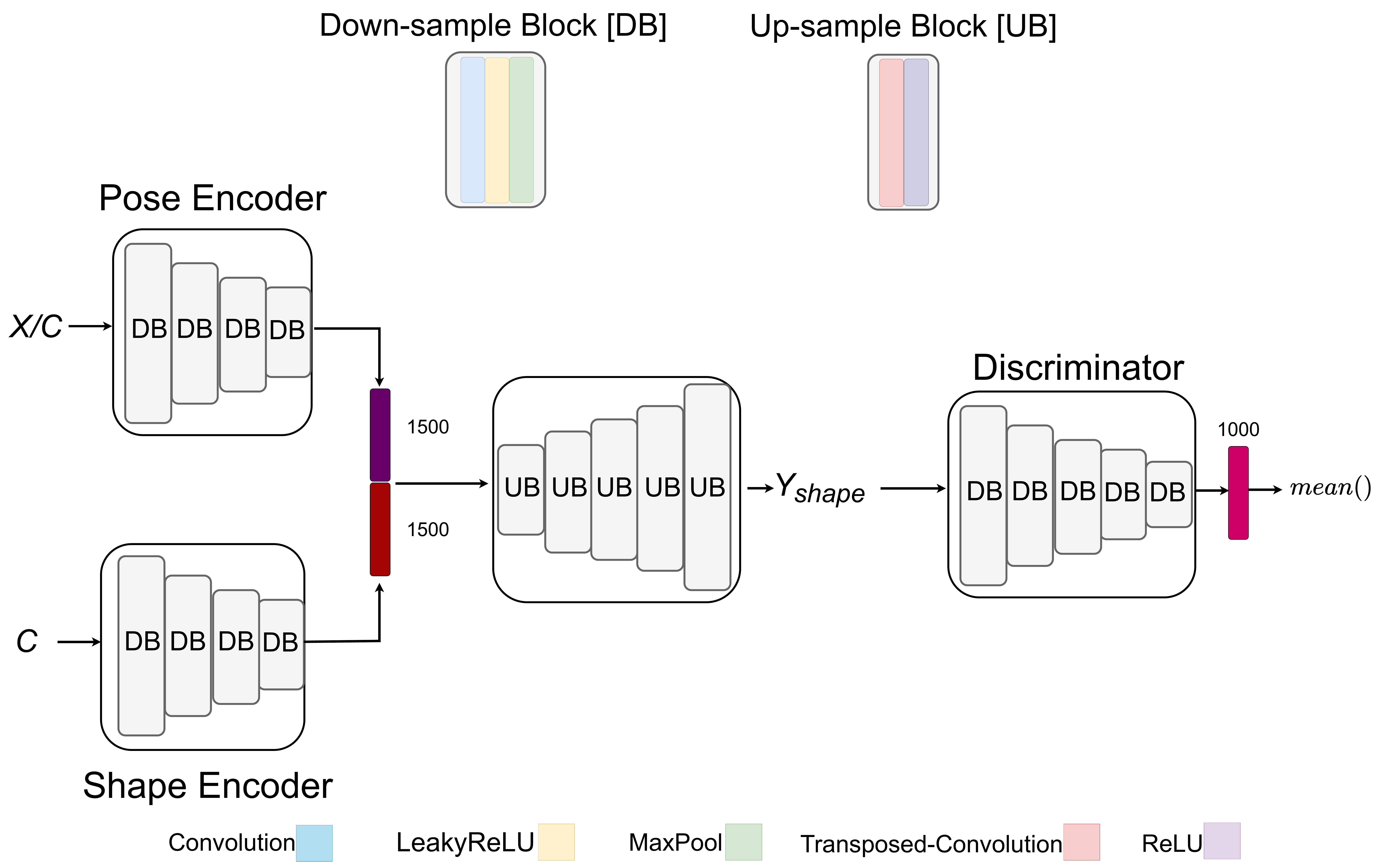} 
\caption{Stage two, in this stage we employ both the original input $X^{d,m}$ and the estimated depth image $Y_{CR}^{d,m}$ to reconstruct 3D shape ($Y_{shape}$).}
\label{fig:reconstruction}
\end{figure}

After reconstructing the default pose, the next stage focuses on 3D volume shape reconstruction, using both the original pose depth image \( X^{d,m} \) and the reconstructed default pose \( C^{d,m} \). The architecture consists of two encoders and a decoder, collectively forming the generator in our GAN framework, as shown in Figure \ref{fig:reconstruction}. Specifically, the pose encoder Eq.~\ref{eq_poseencoder} processes the original input depth image \( X^{d,m} \), while the shape encoder Eq.~\ref{eq_shapeencoder} processes the predicted canonical depth image \( C^{d,m} \). Since the pose may occlude parts of the shape, the shape encoder is designed to avoid obstructions and capture the complete shape for accurate reconstruction. The decoder Eq.~\ref{eq_shape} then reconstructs the high-resolution voxelized shape ($Y_{shape} \in \mathbb{R}^{256 \times 256 \times 256}$).
\begin{equation}
 z_{shape} = ShapeEncoder(Y^d,Y^m)
\label{eq_shapeencoder}
\end{equation}
\begin{equation}
 z_{pose} = PoseEncoder(X^d,X^m)
\label{eq_poseencoder}
\end{equation}
\begin{equation}
 Y_{shape} = recon(X^d,X^m , Y^d,Y^m)
\label{eq_shape}
\end{equation}
To smooth the volume surface \cite{yang2018dense}, we incorporate WGAN-GP in the architecture. We customise the discriminator output to produce a vector rather than a scalar, which stabilises training and improves the quality of the generated surfaces.

\subsection{Loss Function}
The model is divided into two stages during training. \textbf{Stage One} focuses on reconstructing the default pose depth image, while \textbf{Stage Two} is dedicated to reconstructing the original pose in the 3D space.
\newline
\textbf{Stage One}: This stage employs two loss functions: depth loss and mask loss.

\textbf{Depth Loss.} We use Mean Squared Error (MSE) for the depth loss, modified to concentrate on the foreground region.

\begin{equation}
\nonumber L_{Depth} = \frac{1}{N} \sum_{i=1}^N \hat{y}_m y_m(\hat{y}_d - y_d)^2
\label{eq_loss_depth}
\end{equation}

Here, \(\hat{y}_m\) and \(y_m\) denote the predicted mask and the ground truth mask, respectively. Likewise, \(\hat{y}_d\) and \(y_d\) represent the predicted depth and the ground truth depth, respectively. By leveraging the intersection of the masks, we can exclude the background from the depth image, thereby reducing false positive predictions.

\textbf{Mask Loss.} 
For depth image reconstruction, we desire the model to concentrate on the target shape. Consequently, we aim for the model to learn the canonical form mask.

\begin{equation}
 L_{Mask} = \frac{1}{N} \sum_{i=1}^N (\hat{y}_m - y_m)^2
\label{eq_loss_mask}
\end{equation}

\textbf{Combined stage one (SO) loss.} 
Since the model has two objectives, we introduce coefficients \(\alpha\) and \(\beta\) to balance the training.

\begin{equation}
\nonumber L_{SO-weighted} =  \alpha L_{Depth} + \beta L_{Mask}
\label{eq_loss_weighted}
\end{equation}
\newline
\textbf{Stage two}: employs two loss function: reconstruction loss and GAN loss.

\textbf{Reconstruction Loss.} We use Binary Cross-Entropy (BCE) as the loss function. However, empty voxels dominate the reconstruction, leading to false negatives. To address this, we add weights to balance the learning process. The modified BCE is shown below.

\begin{equation}
\nonumber L_{BCE} = -\frac{1}{N} \sum_{i=1}^N [ - \bar{y}_i \log (y_i) - \alpha (1-\bar{y_i}) \log (1-y_i) ]
 \label{modified-bce}.
\end{equation}

The $\alpha$ is the cost weight on the terms, $y$ and $\bar{y}$ are the estimated shape and ground truth shape respectively.

\textbf{GAN loss.} $L_G$ (Eq.~\ref{l_g}) is the loss for the fake estimation, while  $L_D$ (Eq.~\ref{l_d}) is the discriminator loss used by WGAN-GP~\cite{Gulrajani2017}. $y$ represents the reconstructed shape and $\bar{y}$ is the ground truth shape.
In order to tackle vanishing gradients a weight is introduced ($\lambda$) that pushes the gradient norm of the discriminator to be close to~1.

  \begin{equation}
 L_G = -E[D(y|x)]
 \label{l_g}.
 \end{equation}
 
 \begin{equation}
\nonumber  L_D = E[D(y|x)] - E[D(\bar{y}|x)] + \lambda E[(\|\nabla_{\hat{y}} D(\hat{y}|x)\|_2 - 1)^2]
\label{l_d}.
\end{equation}

\textbf{Combined stage two (ST) loss.} As the generator has two objectives, a weight is applied to balance both losses during optimisation as follows:
\begin{equation}
L_{ST-weighted} = \gamma L_{BCE} + (1-\gamma) L_G .
\end{equation}
$L_{ST-weighted}$ is minimised when training the generator, and $L_D$ is minimised when training the discriminator.

\section{Experiments}

\subsection{Training Details}
\label{Training_Details}
We split the non-rigid 
reconstruction task into two stages: \textbf{Stage One}, which focuses on reconstructing the canonical pose of a non-rigid object 
in a depth image, and \textbf{Stage Two}, which reconstructs the volume shape using both the original input depth image and the reconstructed depth image (in the canonical pose).

\textbf{Stage One.} The model was trained for 800 epochs. In the initial phase, specifically for the first 100 epochs, we prioritized mask learning. As mentioned in \ref{Training_Details}, depth images are sensitive to mask intersections; therefore, we set \(\alpha = 10\) and \(\beta = 1000\). In the next 200 epochs, we shifted focus toward the depth objective, setting both \(\alpha\) and \(\beta\) to 1000. For the remaining epochs, we allowed the model to concentrate primarily on the depth objective by setting \(\alpha = 1000\) and \(\beta = 100\). The learning rate was set to 0.001, and we used the Adam optimiser~\cite{Kingma2014}.

\textbf{Stage Two.} The model was trained for 500 epochs. During this stage, we froze the model parameters from Stage One while training the Stage Two model. We set \(\gamma = 0.8\) and the WGAN-GP gradient penalty to \(\lambda = 10\). The learning rate was set to 0.001, and we again used the Adam optimiser~\cite{Kingma2014}.

\subsection{Datasets} \label{dataset}
We conducted our experiments on three datasets, all of which contain non-rigid shapes. Specifically, the dataset from \cite{pickup2016shape} features real human data. This dataset was constructed using the Civilian American and European Surface Anthropometry Resource (CAESAR) \cite{robinette2002caesar}, in which point clouds were fit to templates. In total, it comprises 40 subjects, equally split with 20 males and 20 females. Each subject is represented in 10 different poses.

The second dataset, also from \cite{pickup2016shape}, is a synthetic human dataset. It was created in a parameterised manner using 3D modelling software to control the shape and generate poses. This dataset contains 300 shapes, distributed among 15 subjects: 5 males, 5 females, and 5 children. Each subject has 20 poses.

While the aforementioned datasets focus on humans, real-life scenarios present a variety of non-human, non-rigid subjects. As such, we also chose the TOSCA dataset \cite{bronstein2008numerical}, which includes both humans and animals. In total, the dataset has 80 objects. Due to the varied nature of animals, the numbers of poses differ across objects: two males with 7 and 20 poses respectively; one female with 12 poses; one cat with 11 poses; one dog with 9 poses; one wolf with 3 poses; a horse with 8 poses; a centaur with 6 poses; and one gorilla with 4 poses.

For all datasets, the generation process is as follows: Each shape within the datasets is centred, after which we render an image of size \(500 \times 500\). However, for the TOSCA dataset \cite{bronstein2008numerical}, the sizes of the shapes vary across classes, such as horses and cats. To address this, we scale the shapes 
to a fixed size (bounding box).
We used blender for the dataset generation as we can bind python code to automate the process.

For voxelised ground truth shapes, the datasets offer mesh files which we used for generation.

\section{Evaluation}

\textbf{Stage one}. the evaluation measure is typically based on retrieval results \cite{pickup2018evaluation}. As previous works \cite{pickup2018evaluation} \cite{bronstein2010shrec},  we used Clock Matching and Bag-of-Features (CM-BOF) \cite{lian2013cm} for performing retrieval.
The retrieval task involves ranking the shapes. For each shape in the dataset, we rank the remaining shapes in relation to it. Once ranked, we employ evaluation metrics to assess the retrieval outcomes. From the literature, we adopt four evaluation metrics: Nearest Neighbor (NN), First Tier (FT), Second Tier (ST) and Discounted Cumulative Gain (DCG),   (for more details about CM-BOF or the evaluation metrics, please see the supplementary material).

\textbf{Stage two}. we evaluate our result using Intersection over Union (IOU). The second evaluation metric is mean value Cross-Entropy.
\newline
\textbf{Comparison to prior work.}

\textbf{Stage One}. To the best of our knowledge, there exists no learning-based canonical form model specifically tailored for non-rigid shapes. Consequently, all referenced works herein are non learning-based models.

The majority of the methods mentioned in the literature leverage the Multidimensional Scaling (MDS) technique \cite{elad2003bending}. Hence, (1) MDS-based results are also included in our comparisons. (2) Fast-MDS \cite{faloutsos1995fastmap} projects geodesic distances into a Euclidean space. (3) Non Metric MDS, emphasises preserving the ordering of distances rather than their exact values. (4) Least Squares MDS \cite{elad2003bending}, employs the SMACOF (Scaling by Majorising a Convex Function) algorithm. (5) Constrained MDS \cite{sahilliouglu2015shape} capitalises on the exact correspondence between an original shape and its Landmark MDS embedding. (6) Detail-preserving Mesh Unfolding method \cite{sahilliouglu2016detail} is based on finite elements and omits the use of geodesics. (7) Global Point Signatures (GPS) technique computes the embedding of a mesh. (8) the skeleton based method \cite{pickup2016skeleton} suggests that a skeleton is derived from a mesh to produce a canonical form. For more details about the previous works, please see the supplementary material.

\textbf{Stage Two.} To evaluate our work, we compare it with methods that perform human reconstruction from a single depth image. Few approaches address this specific task: (1) ShapeFormer, presented by \cite{yan2022shapeformer}, estimates shape from partial point clouds. We extracted point clouds from the depth image to run this model. (2) IF-Net, introduced by \cite{chibane2020implicit}, provides two different resolution settings for shape completion experiments. Consequently, we used 300-point clouds (IF-Net:300) and 3000-point clouds (IF-Net:3000) for comparison. The method by \cite{zhou2023human} did not make the code available, 

\begin{figure}[h]
\centering
\includegraphics[width=0.8\textwidth]{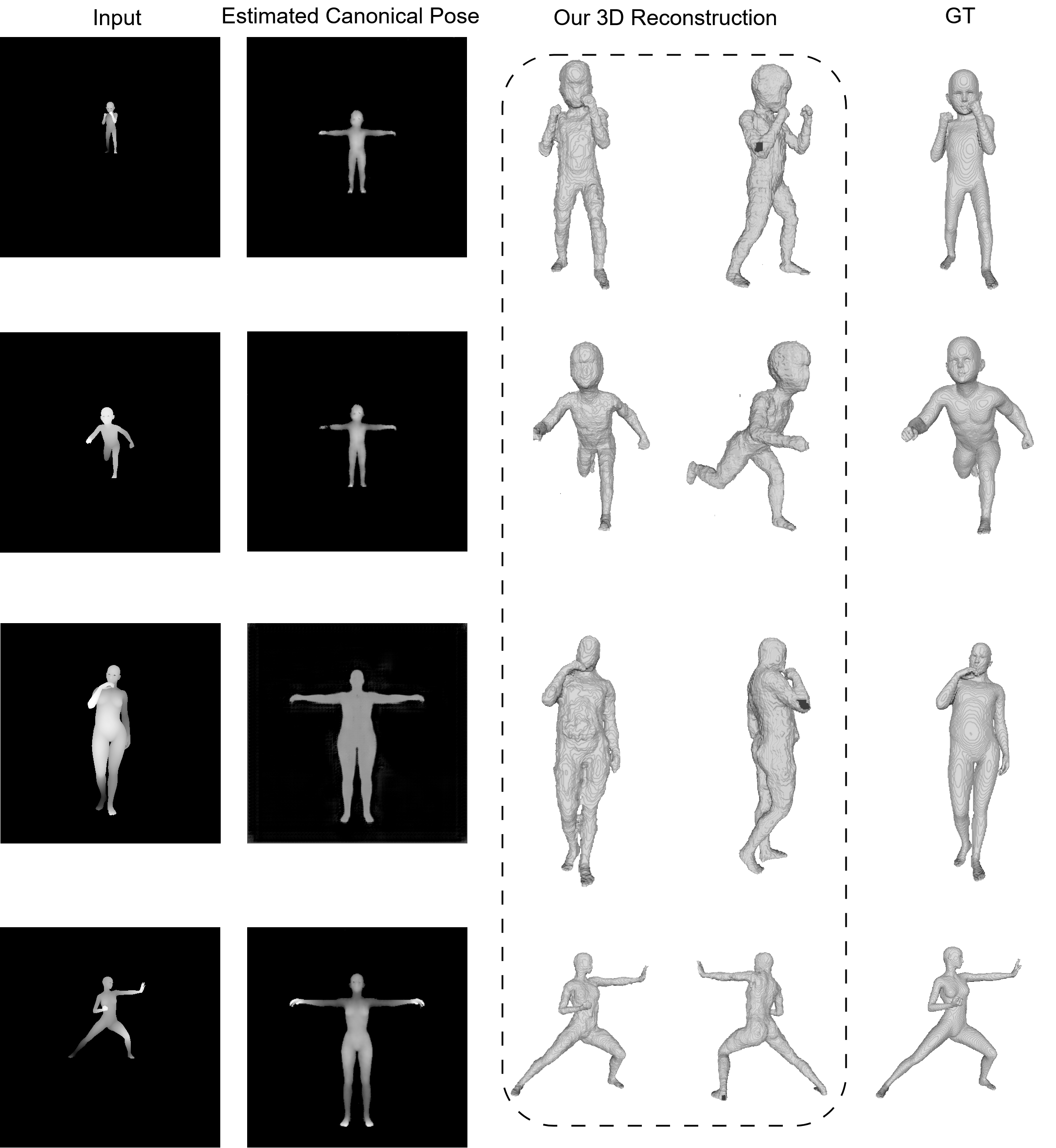} 
\caption{Qualitative results for the synthetic human dataset.}
\label{fig:synthetic_human}

\end{figure}

\begin{figure}[h]
\centering
\includegraphics[width=0.7\textwidth]{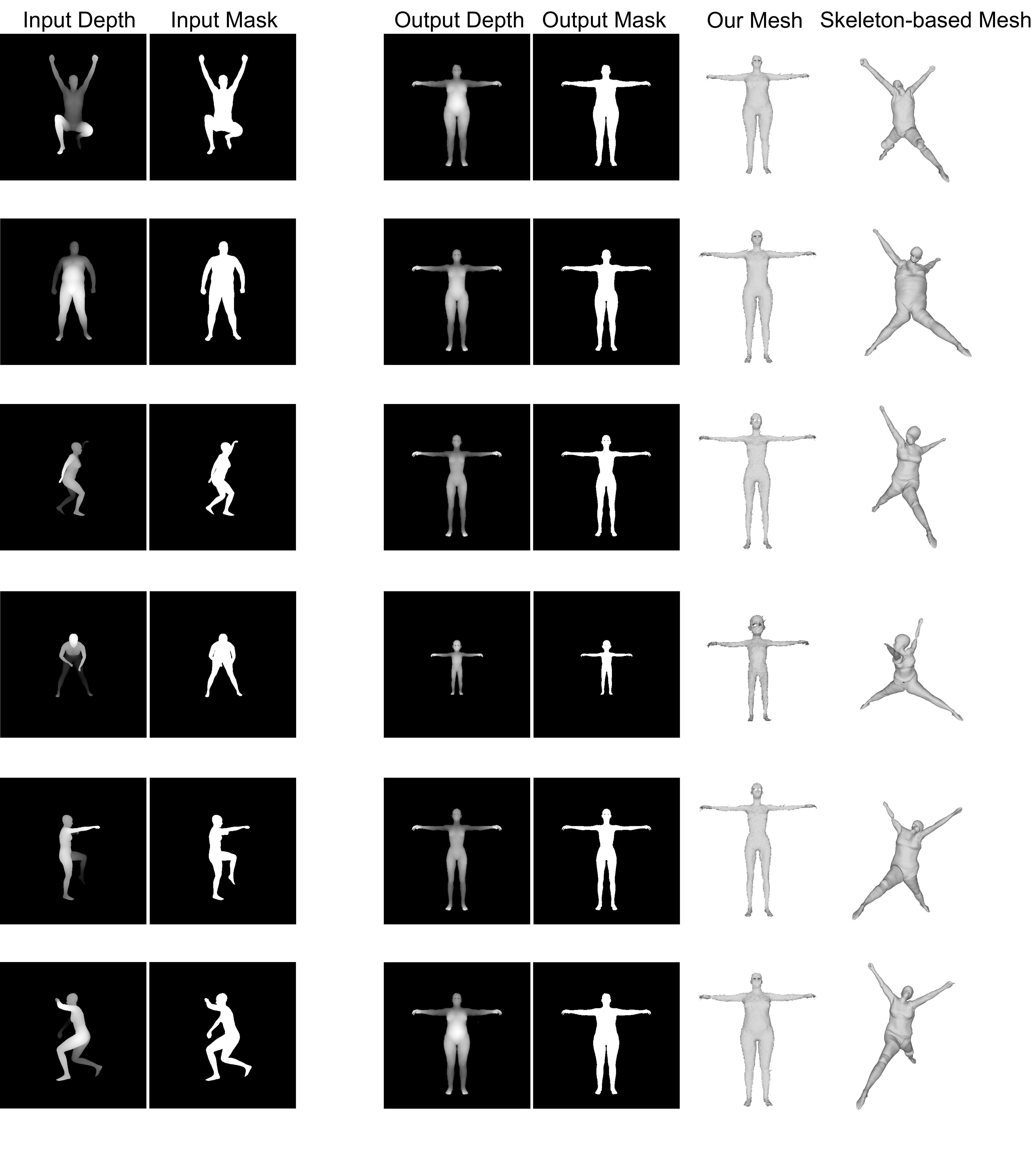} 
\caption{Unseen Canonical form results on real dataset. The model is first trained on synthetic human dataset and then tested on real human dataset. Our meshes are extracted from the output depth images}\vspace{-1mm}
\label{fig:real_human}
\end{figure}

\begin{figure}[h]
\centering
\includegraphics[width=0.6\textwidth]{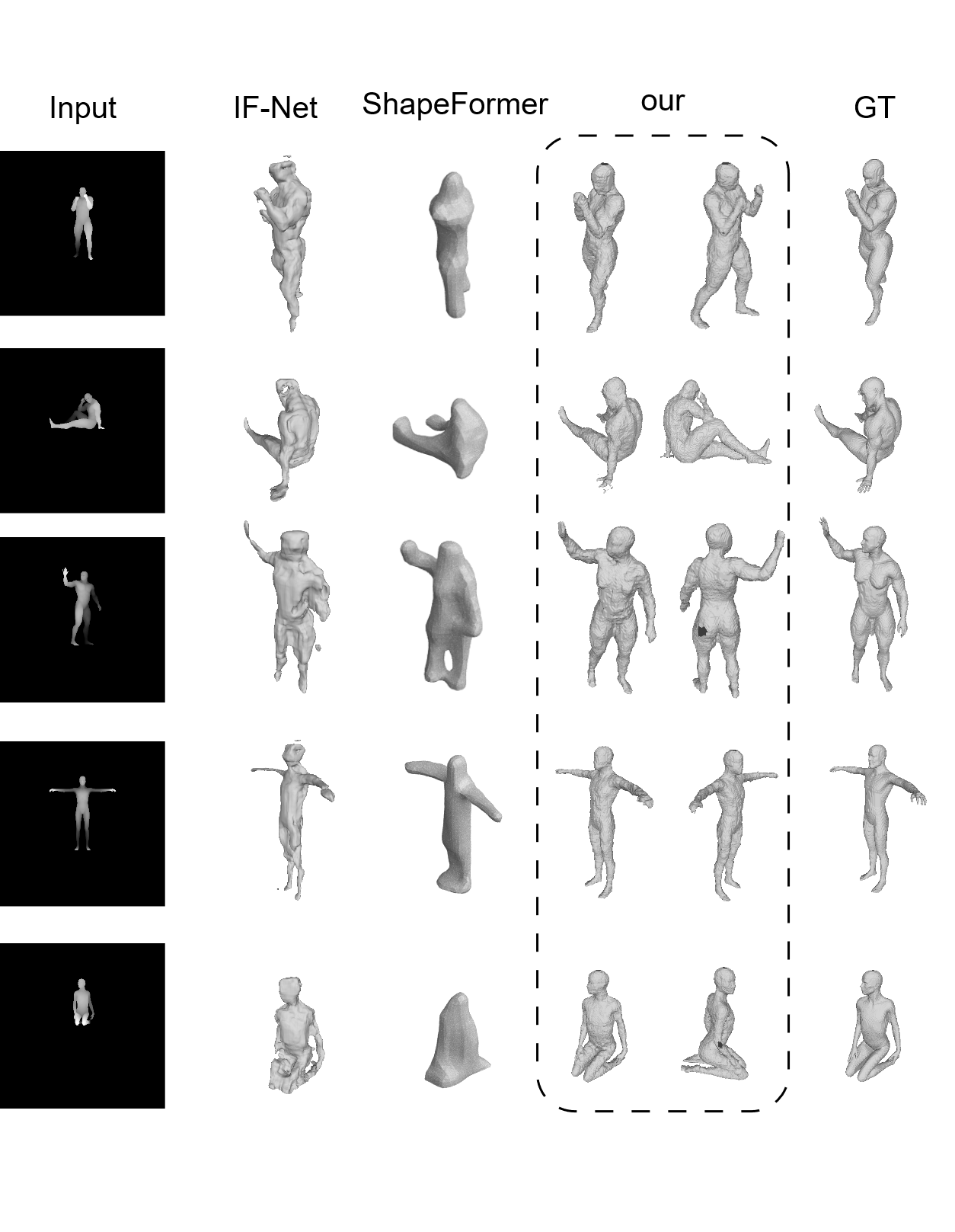} 
\caption{Qualitative results on the synthetic human dataset for Stage two: Pose Recovery in 3D Space.}\vspace{-1mm}
\label{fig:synth_results}
\end{figure}

\begin{figure}[h]
\centering
\includegraphics[width=0.8
\textwidth]{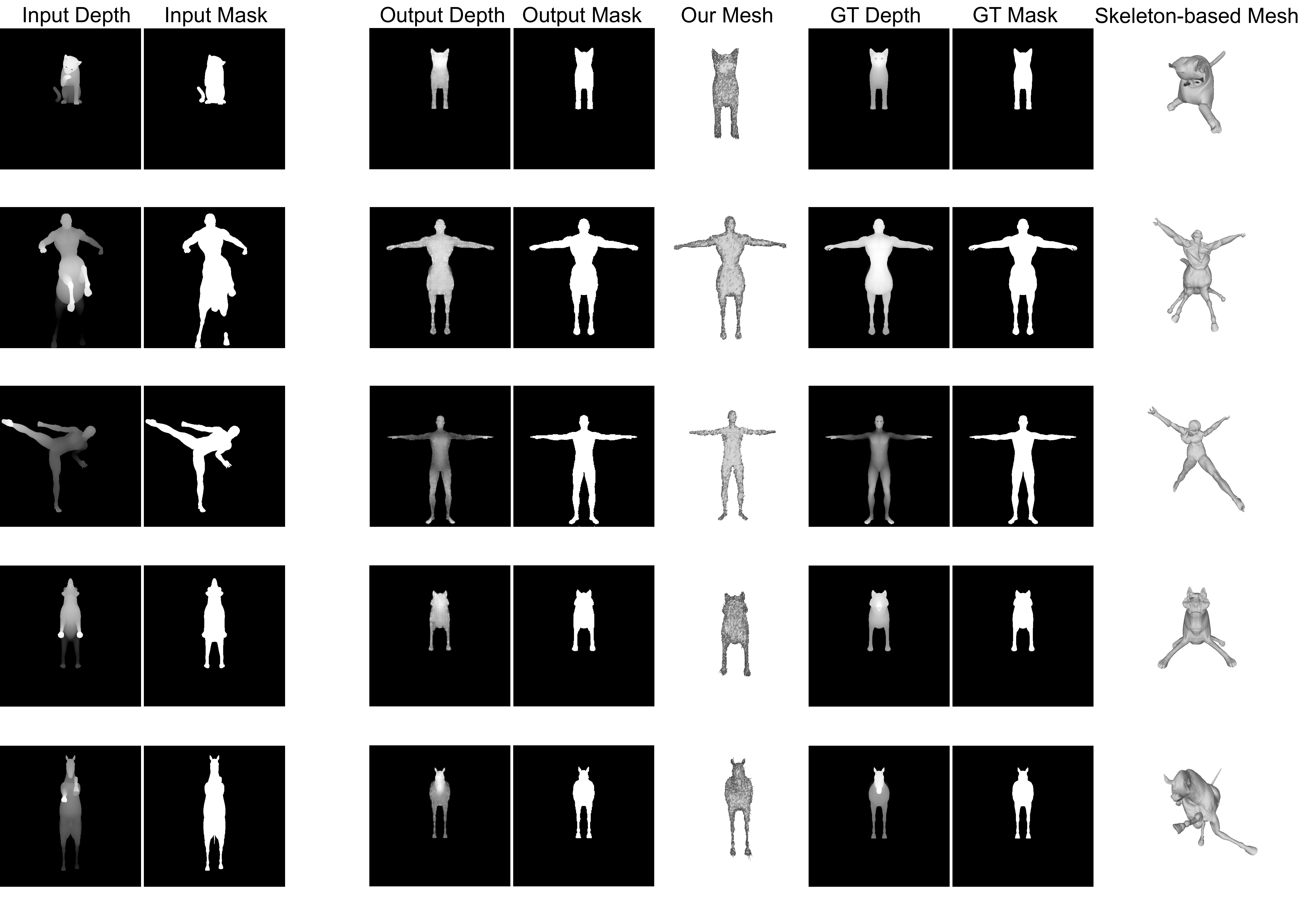} 
\caption{Some canonical form results on the TOSCA dataset. The meshes are extracted from the output depth images}\vspace{-1mm}
\label{fig:tosca}
\end{figure}

\begin{table}[t]
\begin{center}
\caption{Retrieval results for the Synthetic human dataset.} \label{tab:synthetic_human}
\begin{tabular}{|l|c|c|c|c|}
  \hline

    & NN $\uparrow$ & FT $\uparrow$ & ST $\uparrow$ & DCG $\uparrow$
  \\
  \hline  \hline
   
   Classic MDS & 0.10 & 0.22 & 0.39 & 0.54\\
   Fast MDS &0.14 & 0.20 & 0.35 & 0.53\\
   Non-metric MDS &0.09 & 0.24 & \textbf{0.41} & 0.55\\
   Least Square MDS & 0.01 & 0.13 & 0.31 & 0.45\\
   Constrained MDS & 0.04 & 0.14& 0.25& 0.46\\
   GPS & 0.40 & 0.20 & 0.32 & 0.56\\
   Mesh Unfolding & 0.04 & 0.18 & 0.34 &0.49\\
   Skeleton-based & 0.01 &0.14& 0.32 &0.46\\
   Our Method & \textbf{0.51} & \textbf{0.32} & \textbf{0.41} &\textbf{0.63}\\
  \hline
\end{tabular}
\end{center}
\end{table}

\begin{table}[t]
\begin{center}
\caption{Retrieval results for the real human dataset, trained on the synthetic human dataset and tested on the real human dataset} \label{tab:real_human}
\begin{tabular}{|l|c|c|c|c|}
  \hline
  
    & NN $\uparrow$ & FT $\uparrow$ & ST $\uparrow$ & DCG $\uparrow$
  \\
  \hline  \hline
   
   Classic MDS & 0.01& 0.03& 0.07& 0.28 \\
   Fast MDS &0.00 &0.02& 0.04 &0.27\\
   Non-metric MDS &0.02& 0.04& 0.08& 0.30\\
   Least Square MDS & 0.00& 0.00& 0.01& 0.26\\
   Constrained MDS &0.00 &0.01& 0.03& 0.27\\
   GPS &\textbf{0.07} &\textbf{0.06}& \textbf{0.12} &\textbf{0.33}\\
   Mesh Unfolding & 0.00& 0.01 &0.03 &0.28 \\
   Skeleton-based & 0.01 &0.01& 0.02& 0.27\\
   Our Method & 0.04 & 0.023 & 0.051 &0.23\\
  \hline
\end{tabular}
\end{center}
\end{table}

\begin{table}[t]
\begin{center}
\caption{Retrieval results for the TOSCA dataset.} \label{tab:tosca}
\begin{tabular}{|l|c|c|c|c|}
  \hline
    & NN $\uparrow$ & FT $\uparrow$ & ST $\uparrow$ & DCG $\uparrow$
  \\
  \hline  \hline
   
   Classic MDS & 0.74& 0.54& 0.80 &0.80 \\
   Fast MDS &0.73& 0.52& 0.77& 0.77\\
   Non-metric MDS &0.76 &0.67& 0.87& 0.85\\
   Least Square MDS & 0.79& 0.63& 0.86& 0.84 \\
   Constrained MDS &0.88 &0.71&\textbf{0.89}& \textbf{0.89} \\
   GPS &0.71& 0.52& 0.72& 0.76\\
   Mesh Unfolding &0.88 &0.65& 0.86& 0.85 \\
   Skeleton-based & 0.78& 0.62& 0.85& 0.84\\
   Our Method & \textbf{0.91} & \textbf{0.76} & 0.80 &\textbf{0.89}\\
  \hline
\end{tabular}
\end{center}
\end{table}

\begin{table}[t]
\begin{center}
\caption{IoU and Cross entropy evaluation metric for the Synthetic Dataset.} \label{tab:stage_two_synthetic}
\begin{tabular}{|l|c|c|}
  \hline
    & IOU $\uparrow$ & CE $\downarrow$
  \\
  \hline  \hline

   Shapeformer \cite{yan2022shapeformer}  & 0.65 & 0.059\\
   If-net:300\cite{chibane2020implicit} & 0.64 & 0.059 \\
   If-net:3000\cite{chibane2020implicit} & 0.69 & 0.057\\
   Our Method  & \textbf{0.81} & \textbf{0.039}\\
  \hline
\end{tabular} 
\end{center}
\end{table}

\begin{table}[t]
\begin{center}
\caption{IoU and Cross entropy evaluation metric for the TOSCA Dataset.} \label{tab:stage_two_tosca}
\begin{tabular}{|l|c|c|}
  \hline
    & IOU $\uparrow$ & CE $\downarrow$
  \\
  \hline  \hline

   Shapeformer \cite{yan2022shapeformer}  & 0.57 & 0.066\\
   If-net:300\cite{chibane2020implicit} & 0.51 & 0.068 \\
   If-net:3000\cite{chibane2020implicit} & 0.56 & 0.065\\
   Our Method  & \textbf{0.68} & \textbf{0.058}\\
  \hline
\end{tabular} 
\end{center}
\end{table}

\begin{table}[t]
\begin{center}
\caption{Ablation study, IoU and Cross entropy evaluation metric for the TOSCA Dataset.} \label{tab:stage_two_ablation_tosca}
\begin{tabular}{|l|c|c|}
  \hline
    & IOU $\uparrow$ & CE $\downarrow$
  \\
  \hline  \hline

   Complete  & \textbf{0.68} & \textbf{0.058}\\
   w/o Shape Encoder  & 0.59 & 0.067\\
  \hline
\end{tabular} 
\end{center}
\end{table}
\subsection{Results} 

\textbf{Stage One}. Our model is trained on two datasets and tested on three as stated earlier in Section \ref{dataset}. 
For the synthetic human dataset \cite{pickup2016shape}, the results are shown in Table \ref{tab:synthetic_human}. The model is trained using a cross validation method where we perform cross validation across the subjects and poses since poses are similar across the whole subjects. 
Specifically, the subjects and poses are split into groups. Every time, shapes belonging to a chosen group of subjects and a chosen group of poses are used as the test set, while we only use shapes not containing any of these subjects or any of these poses as the training set. This process ensures strict separation of training and test sets during cross validation. The same protocol is applied to other experiments as well.

For the real human dataset \cite{pickup2016shape}, the results are shown in Table~\ref{tab:real_human}. We trained the model on the synthetic dataset and then tested on the real human dataset. Lastly, for the TOSCA dataset  \cite{bronstein2008numerical} the results are shown in Table \ref{tab:tosca}. In the quantitative results, our model outperforms
the state-of-the-art models, except for real human results, 
our result was the second on the NN metric, probably due to the domain gap. 
All the methods performed quite poorly on this dataset, indicating the difficulties for this task.
For the qualitative results, for synthetic human dataset \cite{pickup2016shape}, the results are shown in the supplementary material, and for the real human dataset \cite{pickup2016shape}, the results are shown in Figure \ref{fig:real_human}. For the TOSCA dataset  \cite{bronstein2008numerical} the results are shown in Figure \ref{fig:tosca}.

\textbf{Stage Two.} After training Stage One, we freeze its parameters and train Stage Two. We trained our model on two datasets. For the synthetic human dataset, the results are shown in Table \ref{tab:stage_two_synthetic}. For the TOSCA dataset, the results are shown in Table \ref{tab:stage_two_tosca}. Our model outperforms the state-of-the-art on both datasets. Results for the qualitative results are shown in Figure \ref{fig:synthetic_human} and Figure \ref{fig:synth_results}.

\subsection{Ablation Studies}

In this section, we conduct two ablation studies using the TOSCA dataset, chosen due to its varied content.
Due to space limitations, the qualitative results are presented in the supplementary material.

\textbf{LFE}. Training the model without the LFE component resulted in lower performance compared to the full model. Results are presented in Table \ref{tab:ablation_lfe_mse}.

\textbf{MSFE}. Without the MSFE component, the model's performance was worse compared to the complete model (Table \ref{tab:ablation_lfe_mse}). For classes like dog or cat (which do not have hand or T-pose features), the model could reconstruct the canonical pose. However, for shapes with outstretched hands and legs, such as centaur or human, the results often missed those body parts.

\textbf{Shape Encoder.} As stated earlier in Section \ref{poser_recovery}, the shape encoder helps the model to enhance the reconstruction results. We conducted experiments where the shape encoder was disabled to observe if this led to a degradation in results, the result show in table \ref{tab:stage_two_ablation_tosca}.

\begin{table}[t]
\begin{center}
\caption{Ablation study of LFE and MSFE on the TOSCA dataset.} \label{tab:ablation_lfe_mse}
\begin{tabular}{|l|c|c|c|c|}
  \hline
    & NN $\uparrow$ & FT $\uparrow$ & ST $\uparrow$ & DCG $\uparrow$
  \\
  \hline  \hline

   Complete  & \textbf{0.91} & \textbf{0.76} & \textbf{0.80} &\textbf{0.89}\\
   w/o LFE & 0.88 & 0.62 & 0.76 &0.84\\
   w/o MSFE & 0.72 & 0.48 & 0.61 &0.73\\
  \hline
\end{tabular} 
\end{center}
\end{table}
\section{Conclusion}

In conclusion, our research presents a novel learning-based approach that transforms a single depth image into a standard canonical pose (as a depth image) and then recovers the pose in 3D space. Utilising both a depth image and its associated mask, our model successfully estimates the canonical form even for unseen poses. This approach not only aligns diverse input poses into a unified pose but also extends to accurate shape completion in 3D space. Our method demonstrates robustness in generalizing across varied poses and achieves high fidelity in reconstructing detailed 3D shapes.

\section*{Acknowledgment} 
This work was carried out as part of the Royal Society International Exchange project Computing Canonical Forms from Collections of 3D Models, IES{\textbackslash}R3{\textbackslash}183109.
Additional funding was provided by the Deanship of Scientific Research at Northern Border University, Arar, KSA on project number NBU-FFR-2025-2133-02.

\bibliographystyle{elsarticle-num} 
\bibliography{main}

\end{document}